\documentclass[twoside]{article}
\usepackage{graphics}
\usepackage{graphicx}
\usepackage{amsmath}
\usepackage{amssymb}
\usepackage{booktabs}
\usepackage[numbers]{natbib}
\usepackage{hyperref}
\newcommand{\lecture}[4]{
   \pagestyle{myheadings}
   \thispagestyle{plain}
   \newpage
   \setcounter{page}{1}
   \noindent
   \begin{center}
   \framebox{
      \vbox{\vspace{2mm}
    \hbox to 6.28in { {\bf APMTH 220 Geometric Methods for Machine Learning
                        \hfill Spring 2024} }
       \vspace{4mm}
       \hbox to 6.28in { {\Large \hfill #1  \hfill} }
       \vspace{2mm}
       \hbox to 6.28in { {\it Name: #2 \hfill Submission Date: #3} }
      \vspace{2mm}}
   }
   \end{center}
   \vspace*{4mm}
}

\def\beginrefs{\begin{list}%
        {[\arabic{equation}]}{\usecounter{equation}
         \setlength{\leftmargin}{2.0truecm}\setlength{\labelsep}{0.4truecm}%
         \setlength{\labelwidth}{1.6truecm}}}
\def\endrefs{\end{list}}

\begin{document}

\begin{center}
    {\LARGE \textbf{Graph Attention for Heterogeneous Graphs with Positional Encoding}} \\[1.5em]
    {\large Nikhil Shivakumar Nayak} \\[0.5em]
    {\large Harvard University} \\[0.5em]
    {\large nnayak@g.harvard.edu}
\end{center}

\vspace{1em}

\begin{center}
\textbf{Abstract}
\end{center}

Graph Neural Networks (GNNs) have emerged as the de facto standard for modeling graph data, with attention mechanisms and transformers significantly enhancing their performance on graph-based tasks. Despite these advancements, the performance of GNNs on heterogeneous graphs often remains complex, with networks generally underperforming compared to their homogeneous counterparts. This work benchmarks various GNN architectures to identify the most effective methods for heterogeneous graphs, with a particular focus on node classification and link prediction. Our findings reveal that graph attention networks excel in these tasks. As a main contribution, we explore enhancements to these attention networks by integrating positional encodings for node embeddings. This involves utilizing the full Laplacian spectrum to accurately capture both the relative and absolute positions of each node within the graph, further enhancing performance on downstream tasks such as node classification and link prediction.

\section{Introduction}

Graph Neural Networks (GNNs) have significantly advanced the field of graph-based data analysis, with attention mechanisms \cite{vaswani2017attention} particularly enhancing their performance. Attention in GNNs \cite{velivckovic2017graph} allows the model to focus on the most relevant parts of the graph, adapting the architecture's response based on the input's structure, which is crucial for tasks involving complex node interactions. The efficacy of attention mechanisms has been well-documented across various applications, establishing them as a critical component in modern GNN architectures.

\begin{figure}[h!]
    \centering
    \includegraphics[width=0.8\textwidth]{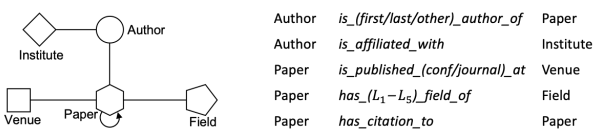}
    \caption{Schema and meta relations of a heterogeneous graph. Figure taken from~\cite{hu2020heterogeneous}.}
    \label{fig:heterogeneous}
\end{figure}

However, the application of GNNs to heterogeneous graphs—graphs with multiple types of nodes and edges—presents a unique set of challenges. Unlike homogeneous graphs, heterogeneous graphs encapsulate a richer semantic structure, making the learning process considerably more complex. This complexity arises from the diverse interactions and different types of relationships that need to be modeled, which traditional GNN approaches struggle to capture effectively. As illustrated in Figure~\ref{fig:heterogeneous}, the Open Academic Graph (OAG) is an example of a complex heterogeneous graph capturing diverse interactions among various entities such as papers, authors, institutions, venues, and fields. This graph exemplifies the meta relations and intricate connectivity typical of academic networks.

Among various GNN architectures, those that incorporate attention mechanisms [\cite{yun2019graph}, \cite{busbridge2019relational}, \cite{hu2020heterogeneous}] have shown promising results even in the context of heterogeneous graphs. This is partly because these models can dynamically learn to emphasize important features and relationships within the graph, adapting to its heterogeneous nature. Our review of benchmark datasets confirms that attention and transformer based models, are particularly effective in tasks like node classification and link prediction within heterogeneous settings.

A pivotal aspect of our research is enhancing the capability of GNNs by incorporating positional encodings \cite{kreuzer2021rethinking} derived from the full Laplacian spectrum of the graph. Positional encodings are crucial in attention mechanisms as they help the model understand the relative and absolute positions of nodes—information that is vital for tasks involving structural and relational reasoning. Traditionally, positional encodings have been utilized in the analysis of homogeneous graphs \cite{kreuzer2021rethinking}, but their potential remains largely untapped in heterogeneous contexts.

In this work, we benchmark various GNN architectures on heterogeneous datasets and enhance the best-performing networks by integrating advanced positional encoding techniques. This approach not only addresses the inherent challenges posed by heterogeneous graphs but also sets a new standard for their analysis, paving the way for more nuanced and effective graph neural network models. A link to the implementation of the code of the paper is available here: \href{https://github.com/NikhilNayak-debug/AM220}{GitHub Repository}.

\section{Background and Notation}

Here we introduce each of the three attention networks that performed the best on the benchmarks and which we will enhance using positional encoding from the Spectral Attention Network (SAN) \cite{kreuzer2021rethinking} that implemented learned positional encoding (LPE) that can take advantage of the full Laplacian spectrum to learn the position of each node in a given graph. In the paper it was done for homogeneous graphs, but now we do it for heterogeneous graphs.

\subsection{Relational Graph Attention Networks (RGAT)}
RGAT \cite{busbridge2019relational} extends the non-relational graph attention mechanisms to handle relational information by integrating multiple relation types into the graph structure. Here, the term "relation type" indicates the type of edge, with nodes having multiple relation types, i.e., edge types. This approach is suitable for complex scenarios like molecular modeling where relations are not uniform.

\textbf{Notation:} $H = [h_1, h_2, \dots, h_N] \in \mathbb{R}^{N \times F}$ represents the feature matrix for $N$ nodes, each with $F$ features. The transformation matrix for relation $r$ is $W^{(r)} \in \mathbb{R}^{F \times F'}$. The attention coefficient $\alpha^{(r)}_{i,j}$ measures the influence of node $j$ on node $i$ under relation $r$.

The key equations include:
\[
G^{(r)} = H W^{(r)} \quad \text{(relation-specific transformation)},
\]
\[
E^{(r)}_{i,j} = a\left( G^{(r)}_i, G^{(r)}_j \right) \quad \text{(computing logits)},
\]
\[
\alpha^{(r)}_{i,j} = \text{softmax}_j(E^{(r)}_{i,j}) \quad \text{(normalizing attention coefficients)},
\]
\[
h'_i = \sigma\left(\sum_{j \in \mathcal{N}^{(r)}_i} \alpha^{(r)}_{i,j} h_j\right) \quad \text{(update rule)},
\]
where $a(\cdot)$ is a learnable function for computing attention coefficients, $\sigma$ is an activation function, and $\mathcal{N}^{(r)}_i$ is the set of neighbors of node $i$ under relation $r$.

\subsection{Graph Transformer Networks (GTN)}

GTNs \cite{yun2019graph} are designed to handle the complexity of heterogeneous graphs by learning and transforming graph structures dynamically. They identify and utilize meta-paths, which are sequences of edges that connect different types of nodes and edges. It applies graph convolution \cite{kipf2016semi} on the learned meta-path graphs, offering a flexible approach to understanding graph structure.

\textbf{Preliminaries:}
\begin{align*}
T_v, T_e &\quad \text{sets of node and edge types}, \\
A_k &\quad \text{adjacency matrix for the } k\text{-th edge type}, \\
X &\in \mathbb{R}^{N \times D} \quad \text{feature matrix for nodes}, \\
R &= t_1 \circ t_2 \circ \ldots \circ t_l \quad \text{composite relation defined by meta-paths}.
\end{align*}

\textbf{Meta-path Adjacency Matrix:}
\[
A_P = A_{t_l} \cdots A_{t_2} A_{t_1} \quad \text{where } P \text{ is a path connecting multiple node types through } t_l \text{ edge types}.
\]

\textbf{Graph Convolution:}
\[
H^{(l+1)} = \sigma\left(D_{\tilde{A}}^{-1/2} \tilde{A} D_{\tilde{A}}^{-1/2} H^{(l)} W^{(l)}\right),
\]
where $\tilde{A} = A + I$ is the adjacency matrix with self-connections, and $D_{\tilde{A}}$ is its degree matrix.

\textbf{Meta-path Generation:}
\[
A^{(l)} = D^{-1} \sum_{t_1, t_2, \ldots, t_l \in T_e} \alpha_{t_1}^{(1)} A_{t_1} \alpha_{t_2}^{(2)} A_{t_2} \ldots \alpha_{t_l}^{(l)} A_{t_l},
\]
where $\alpha_{t_i}^{(i)}$ are learnable weights determining the importance of each edge type at layer $l$.

\subsection{Heterogeneous Graph Transformer (HGT)}

HGT \cite{hu2020heterogeneous} models leverage the unique attributes of heterogeneous graphs by using type-specific parameters for nodes and edges, enhancing the attention mechanism to consider the different types of relationships. The architecture components include Heterogeneous Mutual Attention, Heterogeneous Message Passing, and Target-Specific Aggregation. It utilizes meta relations of heterogeneous graphs to parameterize the weight matrices for each architectural component.

\textbf{Heterogeneous Mutual Attention:}
\[
\text{Attention}_{HGT}(s, e, t) = \text{softmax}\left(\frac{Q(t) K(s)}{\sqrt{d_k}}\right) \cdot W_{\phi(e)},
\]
where $Q(t)$ and $K(s)$ are the query and key projections of nodes $t$ and $s$, and $W_{\phi(e)}$ is a type-specific weight matrix for edge type $\phi(e)$.

\textbf{Message Passing:}
\[
\text{Message}_{HGT}(s, e, t) = \sum_{i=1}^h \text{MSG-head}_i(s, e, t),
\]
where each head computes a part of the message based on type-specific transformations.

\textbf{Target-Specific Aggregation:}
\[
H^{(l+1)}_t = \sum_{s \in \mathcal{N}(t)} \text{Attention}_{HGT}(s, e, t) \cdot \text{Message}_{HGT}(s, e, t),
\]
where $\mathcal{N}(t)$ denotes the neighbors of node $t$.

\subsection{Learned Positional Encoding in Spectral Attention Network (SAN)}
SAN \cite{kreuzer2021rethinking} leverages the entire spectrum of the graph Laplacian to embed positional information into node features, enhancing the model's ability to recognize and differentiate graph structures.

\begin{figure}[h!]
    \centering
    \includegraphics[width=0.95\textwidth]{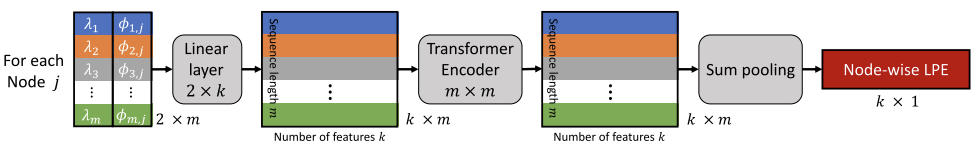}
    \caption{\textbf{Learned positional encoding (LPE) architectures}, with the model being aware of the graph's Laplace spectrum by considering $m$ eigenvalues and eigenvectors, where we permit $m \leq N$, with $N$ denoting the number of nodes. Figure taken from~\cite{kreuzer2021rethinking}.}
    \label{fig:LPE_Architecture}
\end{figure}

\textbf{Learned Positional Encoding (LPE):}
\[
\text{LPE}_j = \sum_{i=1}^m \text{embed}(\lambda_i, \phi_{i,j}),
\]
where $\phi_i$ is the $i$-th eigenvector of the graph Laplacian, $\lambda_i$ is the corresponding eigenvalue, $\phi_{i,j}$ is the $j$-th entry in the $i$-th eigenvector, and $\text{embed}$ is a transformation that combines these values into a positional encoding vector. $\text{LPE}_j$ is the positional encoding for node $j$.

\textbf{Transformer Encoder Layer:}
\[
\text{updated } H^{(l)} = \text{TransformerEncoder}(\text{LPE} \oplus H^{(l)}),
\]
where $\oplus$ denotes concatenation of the learned positional encoding with the node features, and the Transformer Encoder layer applies self-attention mechanisms over the nodes. The LPE architecture is shown in Figure~\ref{fig:LPE_Architecture}. This formulation enables SAN to leverage structural information deeply embedded in the graph's spectrum, providing a robust method for graph analysis tasks.

This background sets the stage for the enhancements we propose, integrating learned positional encodings into these frameworks to address the unique challenges of heterogeneous graphs.

\section{Proposed Approach}

Our study commenced with a systematic benchmarking process to identify optimal network architectures for graph-based tasks, particularly node classification and link prediction. These tasks are central to understanding graph structures, where node classification involves predicting the category of nodes based on graph topology and node features, and link prediction aims to predict potential links between nodes, indicating hidden relationships.

\subsection{Datasets}
We selected diverse datasets to ensure robustness and applicability of our findings across various real-world scenarios:
\begin{enumerate}
  \item \textbf{Molecular Dataset (Tox21):} This dataset is used for predicting the toxicity of chemical compounds, crucial for drug discovery and chemical safety. The node classification task involves predicting whether a compound is toxic. The link prediction task involves predicting potential interactions between different compounds or inferring missing links in chemical reactions.
  \item \textbf{Resource Description Framework (AIFB):} Utilized for semantic web tasks, this dataset involves classifying academic entities, enhancing information retrieval and knowledge discovery in academic databases. In node classification, entities are classified into their respective categories such as researchers, projects, etc. Link prediction tasks involve predicting collaborations.
  \item \textbf{ACM:} Contains metadata from computer science papers, where node classification involves categorizing papers into fields such as Database and Data Mining. Link prediction tasks include predicting co-authorship or citation relationships between papers.
  \item \textbf{IMDB:} Comprises data from the entertainment industry, specifically movies, actors, and directors. Node classification tasks include predicting movie genres. Link prediction involves inferring potential future collaborations between actors.
\end{enumerate}

\subsection{Benchmarking Methodology}
To determine the most effective architectures, we evaluated several networks, spanning from conventional random walk-based methods to state-of-the-art graph neural networks. DeepWalk \cite{perozzi2014deepwalk} applies random walk techniques to generate node embeddings, ignoring node heterogeneity. HERec \cite{shi2018heterogeneous} is a heterogeneous graph embedding method using meta-path based random walks and a skip-gram model. GCN \cite{kipf2016semi}, GAT \cite{velivckovic2017graph} are semi-supervised methods that leverage graph convolution and attention mechanisms, respectively, tested across different meta-paths. HAN \cite{wang2019heterogeneous} utilizes node-level and semantic-level attention mechanisms.

Table \ref{tab:results_transposed} illustrates the performance (F1 score) of various methods on the selected datasets for node classification. Best performances are highlighted in bold.
\begin{table}[h]
  \centering
  \begin{tabular}{l|cccccccc}
    \toprule
    \textbf{Datasets} & \textbf{DeepWalk} & \textbf{HERec} & \textbf{GCN} & \textbf{GAT} & \textbf{HAN} & \textbf{RGAT} & \textbf{GTN} & \textbf{HGT} \\
    \midrule
    Tox21  & 69.50 & 64.70 & 77.20 & 78.30 & 84.50 & 86.40 & 87.00 & \textbf{89.60} \\
    AIFB    & 66.70 & 68.60 & 78.10 & 75.20 & 82.40 & 87.50 & \textbf{90.20} & 89.90 \\
    ACM     & 84.80 & 73.50 & 88.00 & 87.20 & 88.30 & 83.40 & 91.10 & \textbf{93.80} \\
    IMDB    & 50.00 & 47.30 & 51.90 & 52.70 & 52.80 & 70.90 & 71.50 & \textbf{75.10} \\
    \bottomrule
  \end{tabular}
  \caption{F1 scores for various methods on different datasets}
  \label{tab:results_transposed}
\end{table}

The choice of datasets and methods aimed at covering a wide spectrum of graph types and tasks, ensuring that the selected models would be robust and versatile across different scenarios. Specifically, RGAT, GTN, and HGT emerged as the best-performing models, prompting further enhancements using positional encoding strategies tailored for heterogeneous graphs.

\textit{Note:} While the table only shows results for node classification, similar trends were observed for link prediction tasks, with RGAT, GTN, and HGT consistently outperforming other models.

\subsection{Spectral Positional Encoding}
Previous studies have used the eigenfunctions of the Laplacian to provide positional encodings, but the LPE \cite{kreuzer2021rethinking} approach takes this further by exploiting the full expressivity of eigenfunctions. This method offers a principled way to understand graph structures through their spectra. Unlike previous attempts, the approach integrates both absolute and relative positional encodings using eigenfunctions. This provides a nuanced means to measure physical interactions between nodes and to "hear" the structure and sub-structures of the graph.

\begin{figure}[h!]
    \centering
    \includegraphics[width=0.90\textwidth]{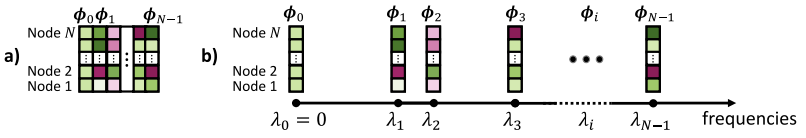}
    \caption{Standard view of the eigenvectors as a matrix on the left and eigenvectors $\phi_i$ viewed as vectors positioned on the axis of frequencies (eigenvalues) on the right. Figure taken from~\cite{kreuzer2021rethinking}.}
    \label{fig:eigenvectors}
\end{figure}

The Laplacian's eigenfunctions, akin to sine functions, enable us to apply the Fourier Transform to graph functions, positioning the eigenvectors on the axis of eigenvalues as illustrated in Figure~\ref{fig:eigenvectors}. This perspective preserves the interpretation of node positions in terms of graph topology. This approach allows for a profound understanding of node relationships through diffusion distances derived from the Laplacian spectrum. Also, these eigenfunctions elucidate the relative positions within the graph.

By leveraging the full spectrum of the graph's Laplacian, our models can effectively distinguish between different graph structures and sub-structures, capturing subtleties that are often missed by other methods. This capability is critical not only for theoretical explorations of graph properties but also for practical applications. Consequently, LPE provides a powerful tool for learning with eigenfunctions, ensuring that our models capture both the geometric and physical properties of graphs.

\subsection{Integration of Spectral Positional Encoding}
Here, we demonstrate how spectral positional encoding is integrated into RGAT, GTN, and HGT architectures. This integration improves their ability to use the graph's structural and spectral features. The positional encoding, derived from the eigenfunctions of the graph Laplacian, is added to the node features before each network processes them with its specific mechanisms.

\textbf{RGAT with Spectral Positional Encoding:}
For RGAT, positional encodings are added to the feature matrix before applying the attention mechanism:
\[
H' = H + \text{LPE},
\]
\[
G^{(r)} = H' W^{(r)},
\]
\[
E^{(r)}_{i,j} = a\left( G^{(r)}_i, G^{(r)}_j \right),
\]

and the rest of the network follows the same equations as before. Here, $\text{LPE}$ is the learned positional encodings, which are added to the node features $H$.

\textbf{GTN with Spectral Positional Encoding:}
In GTN, the positional encoding modifies the graph convolution process by incorporating structural nuances captured by the Laplacian spectrum:
\[
A_P = A_{t_l} \cdots A_{t_2} A_{t_1},
\]
\[
H' = H^{(l)} + \text{LPE},
\]
\[
H^{(l+1)} = \sigma\left(D_{\tilde{A}}^{-1/2} \tilde{A} D_{\tilde{A}}^{-1/2} H' W^{(l)}\right).
\]
The addition of $\text{LPE}$ to $H^{(l)}$ before the convolution operation allows the network to better understand the graph's structural features.

\textbf{HGT with Spectral Positional Encoding:}
In HGT, the integration of \( \text{LPE} \) is used to modify the computation of the attention coefficients, which directly influence the message passing and aggregation processes:
\[
H' = H + \text{LPE} \quad \text{(update node features with positional encoding)},
\]
\[
\text{Attention}_{HGT}(s, e, t) = \text{softmax}\left(\frac{Q(t) K(s, H')}{\sqrt{d_k}}\right) \cdot W_{\phi(e)},
\]
\[
\text{Message}_{HGT}(s, e, t, H') = \sum_{i=1}^h \text{MSG-head}_i(s, e, t, H'),
\]
\[
H^{(l+1)}_t = \sum_{s \in \mathcal{N}(t)} \text{Attention}_{HGT}(s, e, t) \cdot \text{Message}_{HGT}(s, e, t, H').
\]
Here, \( H' \) is used in the key \( K(s, H') \) and message \( \text{Message}_{HGT}(s, e, t, H') \) computations, reflecting how the positional encodings influence the model's understanding of graph structure.

These modifications ensure that each network utilizes the structural context provided by the spectral positional encodings, leading to potentially enhanced performance on graph-based tasks.

\section{Results}

Following a successful theoretical analysis, we integrated spectral positional encoding with state-of-the-art attention networks for heterogeneous graphs. We experimentally evaluate these models on the same tasks and datasets as previously used.

Table \ref{tab:results_performance_improvement} presents the experimental results, showing the absolute mean improvement in F1 score for node classification and link prediction across four datasets:

\begin{table}[h]
\centering
\resizebox{\textwidth}{!}{%
\begin{tabular}{l|cc|cc|cc}
\toprule
\textbf{Dataset} & \multicolumn{2}{c|}{\textbf{RGAT}} & \multicolumn{2}{c|}{\textbf{GTN}} & \multicolumn{2}{c}{\textbf{HGT}} \\
 & \textbf{Node Class.} & \textbf{Link Pred.} & \textbf{Node Class.} & \textbf{Link Pred.} & \textbf{Node Class.} & \textbf{Link Pred.} \\
\midrule
Tox21 & +3, 0.7 & +2, 0.2 & +3, 0.1 & +5, 1.2 & \textbf{+5, 1.3} & \textbf{+5, 1.7} \\
AIFB & -1, 0.2 & +1, 0.5 & +3, 1.1 & \textbf{+5, 1.8} & \textbf{+5, 1.7} & +2, 0.6 \\
ACM & +2, 0.3 & +2, 0.7 & \textbf{+4, 0.9} & \textbf{+3, 0.1} & +1, 0.2 & +2, 0.3 \\
IMDB & +6, 0.9 & +5, 0.7 & +7, 0.9 & +4, 1.2 & \textbf{+8, 2.5} & \textbf{+6, 2.3} \\
\bottomrule
\end{tabular}
}
\caption{Absolute improvement in F1 score (mean, variance) of enhanced models across two tasks and four datasets, averaged over five trials. Bold numbers indicate the highest F1 score achieved (original + improvement) for each dataset/task combination.}
\label{tab:results_performance_improvement}
\end{table}

All experiments were conducted over five trials using different subsets of data, with a 70\%-15\%-15\% train-validation-test split across all datasets. Generally, the integration of spectral positional encoding improved performance. Significant enhancements were particularly noted with the HGT model on the Tox21 and IMDB datasets for both node classification and link prediction, and with the GTN model on the ACM dataset for both node classification and link prediction. A slight but noticeable drop was observed with the RGAT model on the AIFB dataset for node classification, while all models demonstrated mostly improved robust performance across datasets and tasks. Specific declines and major improvements in the models will be further analyzed in the next section. In summary, the spectral positional encoding generally advanced the capabilities of state-of-the-art attention networks on heterogeneous graphs, corroborating our theoretical expectations and computational strategies.

\section{Discussion and Conclusion}

The integration of spectral positional encoding into state-of-the-art attention networks has generally led to improved performance, as evidenced by the results presented in Table \ref{tab:results_performance_improvement}. However, there were specific instances where the results deviated from this trend, notably the slight decrease in F1 score observed with the RGAT model on the AIFB dataset for node classification.

\textbf{Performance Drops:} In the case of RGAT's performance drop in the AIFB dataset, several factors might contribute to this anomaly. The AIFB dataset, primarily used for semantic web tasks, involves complex relational structures that may not be fully captured by the RGAT model even with the addition of spectral positional encoding. Given the dataset's relatively smaller size and higher complexity of semantic relationships, the model may struggle to leverage the additional structural information provided by the spectral encodings effectively. The decrease in performance suggests that the RGAT model may require further tuning or adaptation to better handle the specific characteristics of semantic web data.

\textbf{Significant Improvements:} The notable enhancements observed with the HGT model on the Tox21 and IMDB datasets for both node classification and link prediction, as well as with the GTN model on the ACM dataset, highlight the tailored effectiveness of spectral positional encodings in these contexts. Specifically, the Tox21 dataset, which deals with molecular toxicity prediction, contains intricate molecular interaction patterns. The spectral positional encodings likely improve the model's ability to discern subtle variations in molecular structure that are critical for predicting toxicity. Similarly, in the IMDB dataset, the encodings enhance the model’s capacity to understand complex social networks and collaboration patterns prevalent in the entertainment industry. The ACM dataset, rich in academic collaboration data, benefits from improved recognition of citation patterns and research collaborations.

\textbf{Conclusion:} In this study, we have demonstrated that spectral positional encoding enhances the performance of SOTA attention networks on heterogeneous graphs by leveraging the full geometric information of the graph structure. Learning from eigenfunctions proves crucial for generalizable graph representations, as similar spectra imply similar structures, allowing models to discriminate between nodes and reveal both absolute positions and relative distances. Furthermore, the complexity inherent in heterogeneous graphs—with their varied node and edge types—demands robust modeling approaches such as attention-based frameworks. These frameworks effectively preserve diverse feature information and learn different semantic relations or meta-paths more efficiently. This combination of spectral positional encodings and attention mechanisms addresses common challenges like over-smoothing and over-squashing in message passing neural networks (MPNNs), highlighting the effectiveness of our approach.

\textbf{Future Directions:} Looking ahead, several avenues for further research present themselves based on our findings. First, exploring transformer variants with linear or logarithmic complexity could potentially mitigate the computational bottleneck inherent in the learned positional encoding (LPE), where the complexity scales with the square of the number of eigenvectors \( m \) and the number of nodes \( N \) in the graph. This adjustment could make the approach more scalable. Second, broadening the application of our network to diverse tasks beyond node classification and link prediction could demonstrate the versatility and robustness of our approach in handling a wider range of graph-based problems. Finally, assessing the efficacy of the GTN layer across a range of GNN architectures beyond traditional GCNs could yield improved performance.

\bibliographystyle{unsrt}
\nocite{*}
\bibliography{main}

\begin{thebibliography}{10}

\bibitem{vaswani2017attention}
Ashish Vaswani, Noam Shazeer, Niki Parmar, Jakob Uszkoreit, Llion Jones, Aidan~N Gomez, {\L}ukasz Kaiser, and Illia Polosukhin.
\newblock Attention is all you need.
\newblock {\em Advances in neural information processing systems}, 30, 2017.

\bibitem{velivckovic2017graph}
Petar Veli{\v{c}}kovi{\'c}, Guillem Cucurull, Arantxa Casanova, Adriana Romero, Pietro Lio, and Yoshua Bengio.
\newblock Graph attention networks.
\newblock {\em arXiv preprint arXiv:1710.10903}, 2017.

\bibitem{hu2020heterogeneous}
Ziniu Hu, Yuxiao Dong, Kuansan Wang, and Yizhou Sun.
\newblock Heterogeneous graph transformer.
\newblock In {\em Proceedings of the web conference 2020}, pages 2704--2710, 2020.

\bibitem{yun2019graph}
Seongjun Yun, Minbyul Jeong, Raehyun Kim, Jaewoo Kang, and Hyunwoo~J Kim.
\newblock Graph transformer networks.
\newblock {\em Advances in neural information processing systems}, 32, 2019.

\bibitem{busbridge2019relational}
Dan Busbridge, Dane Sherburn, Pietro Cavallo, and Nils~Y Hammerla.
\newblock Relational graph attention networks.
\newblock {\em arXiv preprint arXiv:1904.05811}, 2019.

\bibitem{kreuzer2021rethinking}
Devin Kreuzer, Dominique Beaini, Will Hamilton, Vincent L{\'e}tourneau, and Prudencio Tossou.
\newblock Rethinking graph transformers with spectral attention.
\newblock {\em Advances in Neural Information Processing Systems}, 34:21618--21629, 2021.

\bibitem{kipf2016semi}
Thomas~N Kipf and Max Welling.
\newblock Semi-supervised classification with graph convolutional networks.
\newblock {\em arXiv preprint arXiv:1609.02907}, 2016.

\bibitem{perozzi2014deepwalk}
Bryan Perozzi, Rami Al-Rfou, and Steven Skiena.
\newblock Deepwalk: Online learning of social representations.
\newblock In {\em Proceedings of the 20th ACM SIGKDD international conference on Knowledge discovery and data mining}, pages 701--710, 2014.

\bibitem{shi2018heterogeneous}
Chuan Shi, Binbin Hu, Wayne~Xin Zhao, and S~Yu Philip.
\newblock Heterogeneous information network embedding for recommendation.
\newblock {\em IEEE transactions on knowledge and data engineering}, 31(2):357--370, 2018.

\bibitem{wang2019heterogeneous}
Xiao Wang, Houye Ji, Chuan Shi, Bai Wang, Yanfang Ye, Peng Cui, and Philip~S Yu.
\newblock Heterogeneous graph attention network.
\newblock In {\em The world wide web conference}, pages 2022--2032, 2019.

\bibitem{li2015gated}
Yujia Li, Daniel Tarlow, Marc Brockschmidt, and Richard Zemel.
\newblock Gated graph sequence neural networks.
\newblock {\em arXiv preprint arXiv:1511.05493}, 2015.

\bibitem{ma2021graph}
Liheng Ma, Reihaneh Rabbany, and Adriana Romero-Soriano.
\newblock Graph attention networks with positional embeddings.
\newblock In {\em Pacific-Asia Conference on Knowledge Discovery and Data Mining}, pages 514--527. Springer, 2021.

\bibitem{hoshen2017vain}
Yedid Hoshen.
\newblock Vain: Attentional multi-agent predictive modeling.
\newblock {\em Advances in neural information processing systems}, 30, 2017.

\bibitem{atwood2016diffusion}
James Atwood and Don Towsley.
\newblock Diffusion-convolutional neural networks.
\newblock {\em Advances in neural information processing systems}, 29, 2016.

\bibitem{defferrard2016convolutional}
Micha{\"e}l Defferrard, Xavier Bresson, and Pierre Vandergheynst.
\newblock Convolutional neural networks on graphs with fast localized spectral filtering.
\newblock {\em Advances in neural information processing systems}, 29, 2016.

\bibitem{rampavsek2022recipe}
Ladislav Ramp{\'a}{\v{s}}ek, Michael Galkin, Vijay~Prakash Dwivedi, Anh~Tuan Luu, Guy Wolf, and Dominique Beaini.
\newblock Recipe for a general, powerful, scalable graph transformer.
\newblock {\em Advances in Neural Information Processing Systems}, 35:14501--14515, 2022.

\bibitem{dwivedi2020generalization}
Vijay~Prakash Dwivedi and Xavier Bresson.
\newblock A generalization of transformer networks to graphs.
\newblock {\em arXiv preprint arXiv:2012.09699}, 2020.

\end{thebibliography}

\appendix
\section{Appendix: Code Implementation}
The implementation of the code used in this paper can be found at the following GitHub repository: \href{https://github.com/NikhilNayak-debug/AM220}{https://github.com/NikhilNayak-debug/AM220}.

\end{document}